# VAPI: Vectorization of Algorithm for Performance Improvement

Mahmood Yashar and Tarik A. Rashid*
Computer Engineering Department. Tishk International University Erbil, Iraq,
Computer Science and Engineering Department, University of Kurdistan Hewler,
mahmood.yashar@tiu.edu.iq
Correspondence Email: tarik.ahmed@ukh.edu.krd

**Abstract**
This study presents the vectorization of metaheuristic algorithms as the first stage of vectorized optimization implementation. Vectorization is a technique for converting an algorithm, which operates on a single value at a time to one that operates on a collection of values at a time to execute rapidly. The vectorization technique also operates by replacing multiple iterations into a single operation, which improves the algorithm's performance in speed and makes the algorithm simpler and easier to be implemented. It is important to optimize the algorithm by implementing the vectorization technique, which improves the program's performance, which requires less time and can run long-running test functions faster, also execute test functions that cannot be implemented in non-vectorized algorithms and reduces iterations and time complexity. Converting to vectorization to operate several values at once and enhance algorithms' speed and efficiency is a solution for long running times and complicated algorithms. The objective of this study is to use the vectorization technique on one of the metaheuristic algorithms and compare the results of the vectorized algorithm with the algorithm which is non-vectorized.

1. INTRODUCTION

Vectorization is a technique that can be applied to algorithms to modify their behavior and enable them to execute multiple values at once. The vectorization-induced change in algorithm behavior results in parallel processing, which reduces the number of iterations, improves algorithm performance in speed, and performs those test functions that cannot be executed in a short amount of time. In time complexity running large algorithms or complex equations in optimal time in real-world applications is very important (Bauckhage, 2018). To apply vectorization technique in Python language provides a standard library to use vectorization this library is known as NumPy. Vectorization plays an important role in different fields of science including field as machine learning, Image processing, and Audio processing to reduce time on processing (Josh Patterson, 2017). Many problems can be solved through optimization in some cases the metaheuristic algorithm performs slow even some algorithms cannot get the result of test functions. The focus of this paper is to apply the vectorization technique to the Ant Nesting Algorithm (ANA), which is a metaheuristic algorithm to improve the algorithm's performance speed.

This study presents a vectorized algorithm to minimize computing time, enhance algorithm speed, make the algorithm clearer to understand, and reduce iterations in the algorithm, which are the major reasons for algorithm speed reduction. Python is a programming language that offers the NumPy library, which can be used to vectorize mathematical calculations, save memory copying, and reduce operation counts. NumPy array is a multinational array that is represented as a block of memory to easily manipulate numbers. It employs a central processing unit (CPU) to work in memory efficiently, storing and accessing multidimensional arrays(Van Der Walt, Colbert, and Varoquaux, 2011). Most programming languages misguidedly use iterations, which results in poor algorithm performance speed, particularly when applied to large datasets. For example, applying one arithmetic operation to an array must use iterations, but in vector representation, the algorithm may perform the same operation on all items without iteration, resulting in increased algorithm performance speed. NumPy arrays provide a compact and powerful way to operate and manipulate data in vectors or matrices (Van Der Walt, Colbert, and Varoquaux, 2011) (Harris *et al.*, 2020).

The objective of this research is to apply vectorization approaches to the ANA metaheuristic algorithm to enhance efficiency, reduce iterations, and improve a test function that does not generate output. Section 2 will present related vectorization work, Section 3 will provide an experiment on the ANA metaheuristic algorithm on how the agents move during each iteration, Section 4 will discuss the mathematical modeling of the ANA algorithm, Section 5 will present the conversion of the ANA algorithm to vectorized version, Section 6 will be based on comparison of speed between vectorized and non-vectorized ANA algorithm finally, Section 7 concludes the topic.

## 2. RELATED WORK

NumPy library has been utilized in several domains such as engineering, economics, finance, chemistry, physics, and astronomy. For example, the NumPy library was used in software for gravitational wave finding (Abbott et al., 2016), also the first imaging of a black hole with the NumPy library. NumPy library is a strong programming concept, which has been used in many scientific researches to improve the performance speed of the algorithms (Chael et al., 2016). NumPy library also improves the speed of the algorithm by improving collaboration reducing error and making the code simpler to understand (Millman and Pérez, 2014). Here are a few examples of how using NumPy's vectorization for speeding up the algorithms in real-world applications:

- Image Processing: NumPy makes it easy to quickly handle images. The basic idea is to use functions that work with collections of coordinates instead of for loops that go through pixel coordinates. Because of this, the answers are ten times faster to adopt than easy ones. In conclusion, NumPy and SciPy make it easy to handle images quickly. By making the code "vectorized," which means that it doesn't use for loops over pixel coordinates but instead uses built-in functions for handling arrays, picture data can be processed quickly in every manner (Bauckhage, 2018).

- Financial Modeling with (NumPy and Pandas): NumPy is an abbreviation for "Numerical Python," and it serves as a basis for many more Python tools. It is mostly used for research computing, particularly data analysis. The purpose of NumPy is to create an array object that is up to 50 times quicker than ordinary Python lists. The NumPy library includes methods for creating nd-arrays with a beginning value that does not need loops. This speeds up the procedure (Sapre and Vartak, 2020).

## 3. EXPERIMENT

To experimentally assess the vectorization outcomes, the ANA algorithm, which is a metaheuristic algorithm, was modified to the vectorization method. In this section, we will examine how the ANA algorithm works, as well as how to vectorize it and compare its results to non-vectorized versions of the algorithms which will be discussed in other sections, this research will examine the vectorization of the ANA algorithm and check the performance of the algorithm after the modification to vectorization. The ANA algorithm is a swarm intelligence metaheuristic algorithm that simulates ant behavior while looking for grains for constructing a new nest. The ANA algorithm analyzes its current and previous positions together with the fitness value to determine the optimal location to lay the grains. It also generates deposition weights using the Pythagorean theorem (Rashid, Rashid, and Mirjalili, 2021). ANA algorithm finds the minimum number among other ants, as a metaheuristic algorithm, the number of agents is initialized to 30 ants with the iteration of 500 and finds the optimal solution among 30 agents, the aim is to reach to minimum number to the destination as shown in the ( Figure 1, Figure 2) all agents try to reach the destination with each iteration each ant move closer to destination.

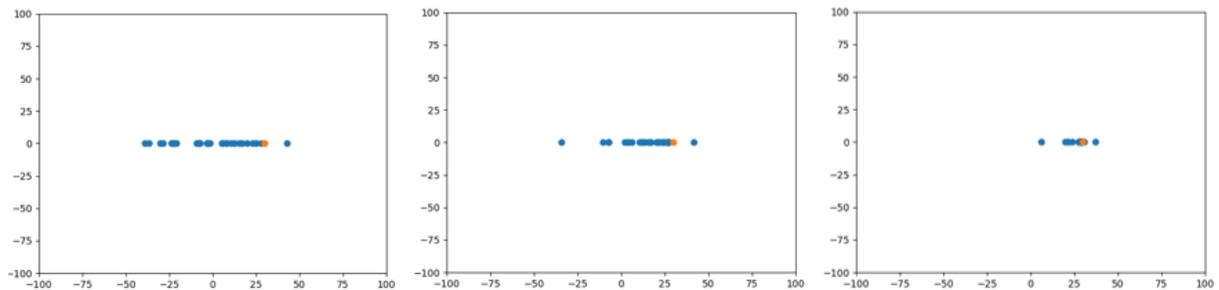

*Figure 1 Agents move to destination from iteration 1 - 250.*

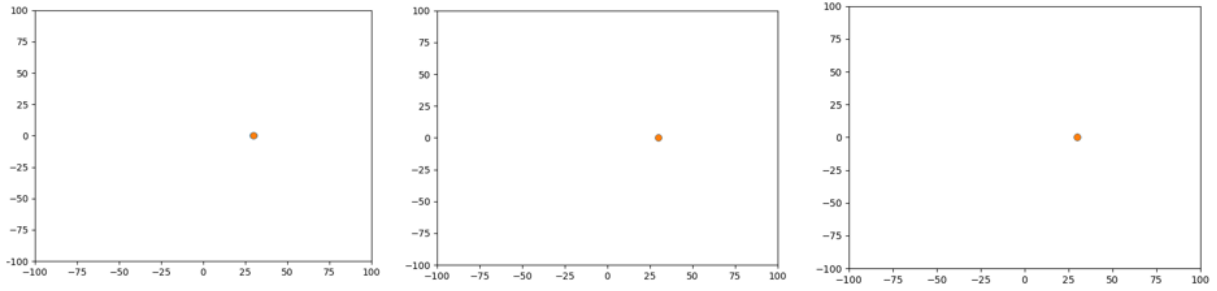

Figure 2 Agents move to destination from iteration 250 - 499.

## 4. Mathematical Modeling of ANA

The ANA algorithm simulates ant behavior to construct a new nest, its significant role is to discover the best solution among other ants. Ants' natural behavior is to gather little grains of grain and transport them to their destinations, as demonstrated in Figure 1, Figure 2. ANA algorithm works to find the minimal distance, which is used to put the grains close together to construct a strong wall around the queen ant. Firstly, the algorithm starts by initializing ten random numbers between [lower boundary to upper boundary] as [-100, 100] for each ant, to determine which position to put the grains in and find the best position. In this case, the new position will be compared with the old position to determine which one is the optimal position. The worker ant starts searching for a new position randomly which the new position is stated as (Xt,i+1) which represents the new deposition position, and (Xt, i) represents the current position, and the deposition rate of change is denoted as (ΔXt,i+1) as shown in equation (1).

$$x_{t,i+1} = x_{t,i} + \Delta x_{t,i+1} \quad\quad 1$$

(ΔXt,i+1) represents the rate of change of the deposition position which depends on the (dw) deposition position multiplied by the difference between the best worker ant, which is represented as (Xt,i) best, and the current deposition position presented as (Xt,i) as shown in equation (2).

$$\Delta x_{t,i+1} = dw \times (x_{t,i\ best} - x_{t,i}) \quad\quad 2$$

As the algorithm starts there is no best ant it will choose the first ant as the best ant then the comparison begins. The fitness function for each ant will be determined for all 30 ants, then the best fitness will be selected, and two new arrays will be created for a new position and previous position after it will compare every 10 random numbers for each ant between the first and last ant to find the distance between the best ant, distance is subtracting the best ant with first ant and it will create the previous variable and create simple random walk between [-1,1]. Then the algorithm will pass through three conditions.

A. If the current ant equals the best ant, it will use equation (3).

$$\Delta x_{t,i+1} = r \times x_{t,i} \quad\quad 3$$

B. If the current ant equals the previous ant, it will use equation (4).

$$\Delta x_{t,i+1} = r \times (x_{t,i\ best} - x_{t,i}) \quad\quad 4$$

C. Else use equation (5).

$$dw = r \times (\frac{T}{T_{previous}})  \qquad 5$$

(dw) represents the deposition weight and r is the random number between [-1,1], also the algorithm uses levy flight to move to the algorithm more suitable (Jain, Sharma, and Sharma, 2021). Also, the (dw) depends on the (T) which known tendency rate using the Pythagorean theorem and (T previous) previous tendency rate as shown in equations (6 and 7). The tendency rate (T) is represented in equation (6).

$$T = \sqrt{(X_{t,ibest} - X_{t,i\,previous})^2 + (X_{t,ibestfitness} - X_{t,ipreviousfitness})^2} \qquad 6$$

The previous tendency of worker ants represented in equation (7)

$$Tprevious = \sqrt{(X_{t,ibest} - X_{t,i\,previous})^2 + (X_{t,ibestfitness} - X_{t,ipreviousfitness})^2} \qquad 7$$

Equations 6 and 7 in the ANA algorithm have errors, contrary to what is given in the Pythagorean theorem equation. The errors, along with the proper calculations, may be found in equations 6 and 7. The preceding piece of work had a typographical error, the incorrectly calculated equation would not provide the right output for the algorithm, the error in the equation was that it used subtraction instead of addition (Rashid, Rashid, and Mirjalili, 2021). The corrected equation for the Pythagorean theorem has been applied to the ANA algorithm. In addition, the corrected equation has been implemented in this research. The VANA (Vectorized Ant Nesting Algorithm) uses the correct version of the equation, and all of the output of the VANA algorithm uses the correct equation of the Pythagorean theorem. This is also stated in the pseudocode of the algorithm.

According to the Pythagorean theorem, the square of the length of the hypotenuse in a triangle with a right angle is equal to the sum of the squares of the lengths of the other two sides of the triangle. The hypotenuse is the side that is opposite the right angle. It is possible to describe it mathematically using the formula a2 + b2 = c2, where a and b refer to the lengths of the two legs (the sides next to the right angle), and c refers to the length of the hypotenuse. Pythagoras, an ancient Greek mathematician, is given credit for both the discovery of this theorem and its proof. Consequently, this theorem bears his name. It finds use in a wide variety of disciplines, including geometry, trigonometry, and physics, among others. This optimization algorithm also uses the Pythagorean theorem to identify the best position with the help of this equation. The ANA algorithm uses this equation to help find the optimum solution by referring to the tendency rate and previous tendency rate as stated in the equation (6, 7). Finally, the Pythagorean theorem has been implemented in the algorithm to find the optimum solution. Improving the speed of an algorithm is important in a variety of scientific fields, and this article will discuss how the ANA algorithm may be vectorized to increase its processing speed. This will allow the algorithm to do several operations on the data being processed simultaneously, rather than just one. Figure 3 represents the pseudocode of the ANA algorithm.

```
Initialize worker ant population X_i (i = 1, 2, 3, ..., n)
Initialize worker ant previous position X_{iprevious}
while iteration (t) limit not reached
   for each artificial worker ant X_{t,i}
      find best artificial worker ant X_{t,ibest}
      generate random walk r in [-1, 1] range
      if (X_{t,i} == X_{t,ibest})
         calculate ΔX_{t+1,i} using Equation (3)
      else if (X_{t,i} = X_{t,iprevious})
         calculate ΔX_{t+1,i} using Equation (4)
      else
         calculate T using Equation (6)
         calculate T_{previous} using Equation (7)
         calculate dw using Equation (5)         // for minimization
         calculate ΔX_{t+1,i} using Equation (2)
      end if
         calculate X_{t+1,i} using Equation (1)
      if (X_{t+1,i} fitness < X_{t,i} fitness)    // for minimization
         move accepted and X_{t,i} assigned to X_{t,iprevious}
      else
         maintain current position
      end if
   end for
end while
```

*Figure 3 ANA Algorithm pseudocode (Rashid, Rashid, and Mirjalili, 2021).*

Some of the test functions were significantly slow when performing the ANA algorithm, as demonstrated in (Table 1, and Table 2) It suggested that (F9, F11, F12, F15, F17, and F18) and primarily CEC01 and CEC06 are likewise very slow (Rashid, Rashid and Mirjalili, 2021) Because of the slowness of CEC01 and CEC06, their output was not used in the ANA algorithm which priorities to the slow performance of the algorithm as shown in Table 3. After using the vectorization approach, it increases the speed of the algorithm and receives CEC01 and CEC06 output. The explanation for the algorithm's slow performance is clarified in the next session.

*Table 1 Time in seconds for F1 to F18 test functions.*

| Test Function | Execution Time ANA Algorithm |
|---|---|
| F1 | 32 Seconds |
| F2 | 57 Seconds |
| F3 | 52 Seconds |
| F4 | 28 Seconds |
| F5 | 26 Seconds |
| F7 | 26 Seconds |
| F9 | **106 Seconds** |
| F10 | 16 Seconds |
| F11 | **168 Seconds** |
| F12 | **140 Seconds** |
| F13 | 6 Seconds |
| F14 | 39 Seconds |
| F15 | **151 Seconds** |
| F16 | 79 Seconds |
| F17 | **81 Seconds** |
| F18 | **96 Seconds** |

*Table 2 Time in seconds for CEC01 to CEC10 test functions.*

| Test Function | Execution Time ANA Algorithm |
|---|---|
| CEC01 | **23705 Seconds ,6.58472222(Hours)** |
| CEC02 | 58 Seconds |
| CEC03 | 511 Seconds |
| CEC04 | 208 Seconds |
| CEC05 | 337 Seconds |
| CEC06 | **10606 Seconds, 2.94611111 (Hours)** |
| CEC07 | 341 Seconds |
| CEC08 | 266 Seconds |
| CEC09 | 148 Seconds |
| CEC10 | 17 Seconds |

Table 3 CEC01 and CEC06 haven't been recorded (Rashid, Rashid, and Mirjalili, 2021).

| Test Function | ANA | |
| --- | --- | --- |
| | Mean | Standard Deviation |
| CEC01 | - | - |
| CEC02 | 4 | 2.87 x 10-14 |
| CEC03 | 13.70240422 | 2.01 x 10-11 |
| CEC04 | 38.50887822 | 10.07245727 |
| CEC05 | 1.224598709 | 0.114632394 |
| CEC06 | - | - |
| CEC07 | 116.5962143 | 8.825046006 |
| CEC08 | 5.472814997 | 0.429461877 |
| CEC09 | 2.000963996 | 0.00341781 |
| CEC10 | 2.718281828 | 4.4 x 10-16 |

## 5. Vectorization of ANA Algorithm

As previously mentioned, the ANA Algorithm is extremely slow because of a large number of iterations since it starts by initializing the ants with 10 random numbers between [lower boundary to upper boundary] as [-100, 100] for each ant, which it will cycle 300 times, however, the NumPy function can handle this looping issue with its library. as shown in Figure 4.

```python
for agent, _ in enumerate(range(self.number_of_agents)):
    x = []
    for _ in range(self.dimenssion):
        x.append(self.get_random_xy())
    ant = Ant.Ant(x, agent, Functions, x, self.function)
    self.worker_ants.append(ant)
```

Figure 4 Initializing Random number of [lower boundary, upper boundary] with for loops.

Instead of dealing with loops to repeat 300 times to initialize 10 random numbers between [lower boundary, upper boundary] as [-100, 100] for each ant, the NumPy module can generate a size of (dimensions, agents) matrix, that the process can be vectorized to enhance its performance of the algorithm, as shown in Figure 5.

```python
self.worker_ants = np.random.uniform(self.lower_bound,self.upper_bound,(self.dimenssion,self.number_of_agents))
```

Figure 5 NumPy array for creating a matrix of (dimensions, agents) random number between [lower boundary, upper boundary].

Following the execution of the np.random function, as shown in Figure 5, a matrix of (dimensions, number of ants) size will be created. As shown in Figure 6 the rows represent the dimensions of ants, and the column represents the number of each ant.

$$\text{workerAnts} = \begin{bmatrix} \text{Ant}_1 & \text{Ant}_2 & \ldots & \ldots & \text{Ant}_{n-1} & \text{Ant}_n & \\ A_1, D_1 & A_2, D_1 & \ldots & \ldots & A_{n-1}, D_1 & A_n, D_1 & \text{Dimention}_1 \\ A_1, D_2 & A_2, D_2 & \ldots & \ldots & A_{n-1}, D_2 & A_n, D_2 & \text{Dimention}_2 \\ \ldots & \ldots & \ldots & \ldots & \ldots & \ldots & \\ \ldots & \ldots & \ldots & \ldots & \ldots & \ldots & \\ A_1, D_{n-1} & A_2, D_{n-1} & \ldots & \ldots & A_{n-1}, D_{n-1} & A_{n-1}, D_{n-1} & \text{Dimention}_{n-1} \\ A_1, D_n & A_2, D_n & \ldots & \ldots & A_{n-1}, D_n & A_n, D_n & \text{Dimention}_n \end{bmatrix}$$

Figure 6 Current view of worker ant of (dimensions, agents) matrix.

The algorithm requires a previous position after initializing the ants. If there is no previous ant in the first step, the current worker ant will be initialized as a previous one. To transfer all current worker ants to previous worker ants, no iteration is required in the NumPy array otherwise it will cause the deduction of the speed of the algorithm. The previous position will be used to keep track of the current position and newly generated random value to know which value is more optimum NumPy arrays contain a copy library that eliminates the need for iterative processes which will copy all current ant to the previous ant, as shown in Figure 7 and Figure 8.

```
previous_Ant = np.copy(self.worker_ants)
```

*Figure 7 Copying the current ant to the previous ant.*

$$\text{Previous Ant} = \begin{bmatrix} A_1, D_1 & A_2, D_1 & \ldots & \ldots & A_{n-1}, D_1 & A_n, D_1 \\ A_1, D_2 & A_2, D_2 & \ldots & \ldots & A_{n-1}, D_2 & A_n, D_2 \\ \ldots & \ldots & \ldots & \ldots & \ldots & \ldots \\ \ldots & \ldots & \ldots & \ldots & \ldots & \ldots \\ A_1, D_{n-1} & A_2, D_{n-1} & \ldots & \ldots & A_{n-1}, D_{n-1} & A_n, D_{n-1} \\ A_1, D_n & A_2, D_n & \ldots & \ldots & A_{n-1}, D_n & A_n, D_n \end{bmatrix} \begin{matrix} Dimention_1 \\ Dimention_2 \\ \\ \\ Dimention_{n-1} \\ Dimention_n \end{matrix}$$

with columns $Ant_1, Ant_2, \ldots, Ant_{n-1}, Ant_n$ (Copy of the worker ant).

*Figure 8 Current view of previous ants.*

The rate of change it will create (dimensions, agents) shape with zero values as shown in Figure 9, Figure 10.

```
rate_of_change = np.zeros_like(self.worker_ants)
```

*Figure 9 Matrix of (dimensions, agents) zeros.*

$$\mathbf{RateOfChange} = \begin{bmatrix} 0 & 0 & \ldots & \ldots & 0 & 0 \\ 0 & 0 & \ldots & \ldots & 0 & 0 \\ \ldots & \ldots & \ldots & \ldots & \ldots & \ldots \\ \ldots & \ldots & \ldots & \ldots & \ldots & \ldots \\ 0 & 0 & \ldots & \ldots & 0 & 0 \\ 0 & 0 & \ldots & \ldots & 0 & 0 \end{bmatrix} \begin{matrix} Dimention_1 \\ Dimention_2 \\ \\ \\ Dimention_{n-1} \\ Dimention_n \end{matrix}$$

with columns $Ant_1, Ant_2, \ldots, Ant_{n-1}, Ant_n$.

*Figure 10 Current view of (dimensions, agents) matrix of zeros.*

```
worker_ant_fintness = self.get_fitness(self.function, self.worker_ants)
```

*Figure 11 Worker ant fitness finds fitness of ants by creating (1,30) matrix.*

Worker ant fitness as stated in Figure 11 will find the fitness of all ants according to the test function and store them in a (1,30) shape matrix as shown in Figure 12.

$$\textbf{WorkerAntFit} = \begin{bmatrix} \overset{\text{Ant}_1}{Fitness of ant_1} & \overset{\text{Ant}_2}{Fitness of ant_2} & \ldots & \ldots & \overset{\text{Ant}_{n-1}}{Fitness of ant_{n-1}} & \overset{\text{Ant}_n}{Fitness of ant_n} \end{bmatrix} \text{Dimention}_1$$

Figure 12 Worker ant fitness contains fitness of all ants with the shape of (1,30) matrix.

```
previous_fitness = self.get_fitness(self.function, previous_x)
```

Figure 13 Previous fitness stores the fitness of the previous ant with (1,30) shape matrix.

ANA algorithm will also find the previous ant fitness to keep track of which ant has optimal fitness value, same as worker ant fitness with the shape of (1,30) as shown in Figure 13, Figure 14.

$$\textbf{PrevAntFit} = \begin{bmatrix} \overset{\text{Ant}_1}{PrevAntFit_1} & \overset{\text{Ant}_2}{PrevAntFit_2} & \ldots & \ldots & \overset{\text{Ant}_{n-1}}{PrevAntFit_{n-1}} & \overset{\text{Ant}_n}{PrevAntFit_n} \end{bmatrix} \text{Dimention}_1$$

Figure 14 Current view of previous ant fitness.

The best ant index stores the index of the best ant fitness case if the best fitness is the first ant it will store the index of 0 and the best ant fitness will store the fitness of the best ant as shown in Figure 15 and Figure 16.

```
best_ant_index, best_ant_fitness = self.get_best_ant(worker_ant_fintness)
```

Figure 15 Index stores the index of best ant.

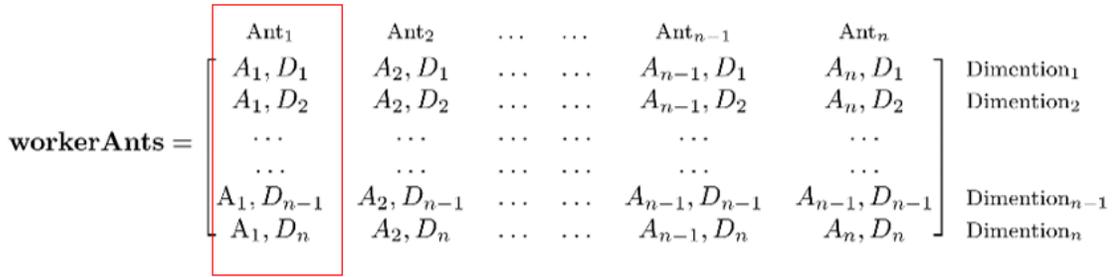

Figure 16 Current view of best ant fitness with index.

Best ant stores the worker ant best ant which indicates that select all column of best ant index and reshape it according to the dimension of the fitness function, index is mainly used to indicate the position of the best ant and[:] double colon represents all column and the best ant index is the index of the best ant as shown in Figure 17, Figure 18.

```
best_ant = self.worker_ants[:, best_ant_index].reshape((self.dimenssion, 1))
```

Figure 17 Setting best ant with best ant index.

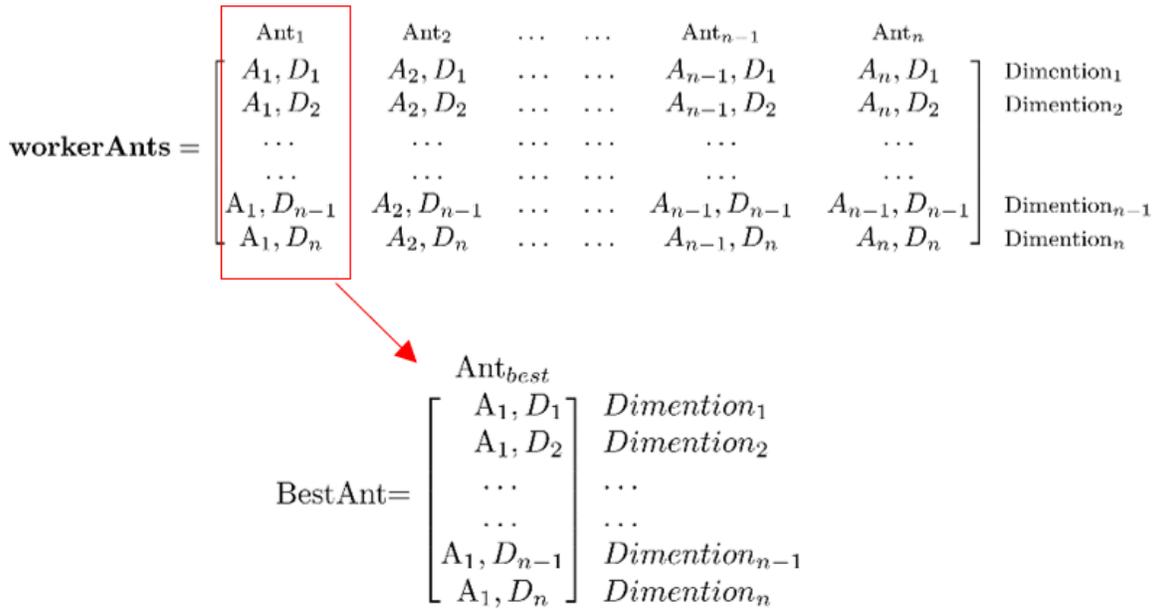

Figure 18 Current view for identifying best ant.

Distance from best ant stores all matrix of worker ant (10,30) then subtracts the best ant by all other ants this process of subtracting the ants with the best ant is to find the distance between the best ant and other ants as shown in Figure 19, Figure 20.

```
distance_from_best_ant = best_ant - self.worker_ants
```

Figure 19 Identifying the distance of other ants with the best ant.

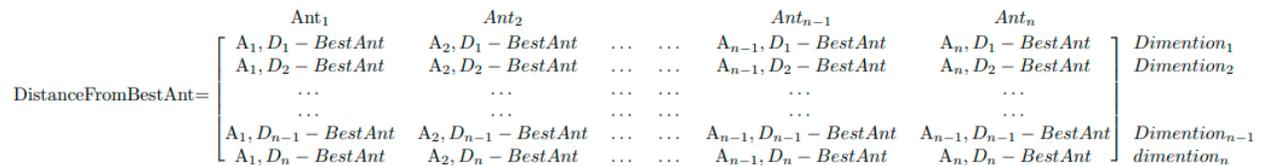

Figure 20 Current view of subtracting the best ant from all other ants.

Temp_xs with the help of the NumPy library will copy all worker ant to the temp_xs without using iteration as shown in Figure 21, Figure 22, and Figure 23.

```
temp_xs = np.copy(self.worker_ants)
```

Figure 21 Copying worker ant to temp_xs.

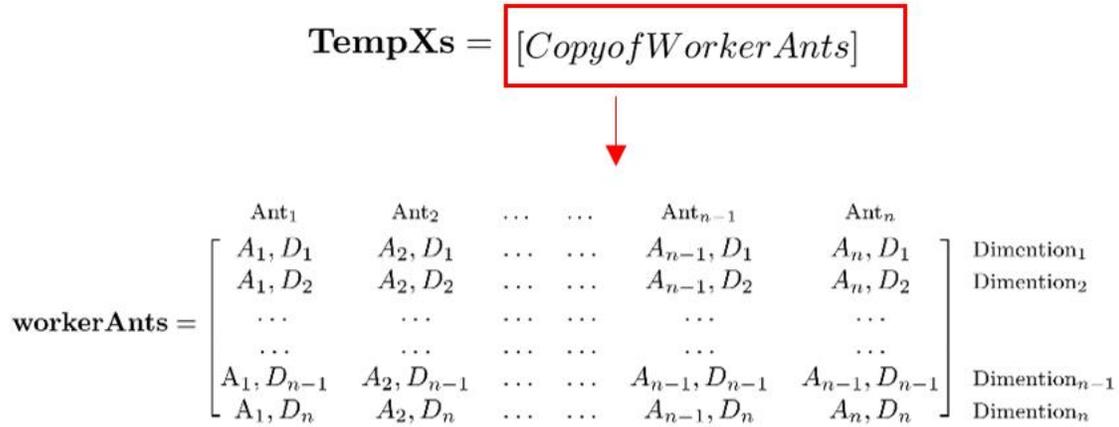

Figure 22 Current view of copying worker ants to temp_xs.

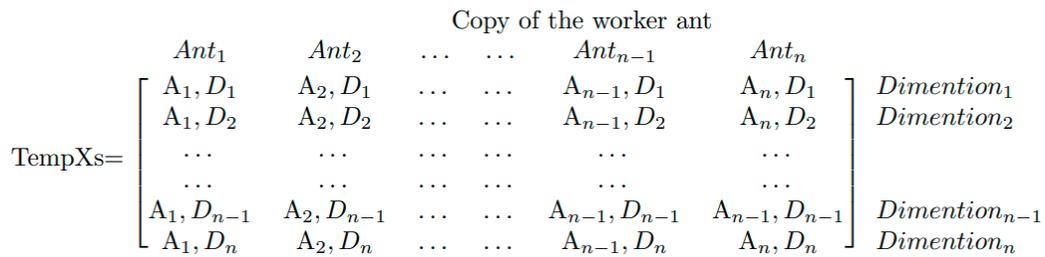

Figure 23 Current view of worker ant to temp_xs.

The NumPy library also provides a matrix initialization of ones with the same dimension of worker ant as stated in Figure 24 and shown in Figure 25.

```
r = np.ones_like(self.worker_ants)
```

Figure 24 Initializing matrix of 1 with the size of (10,30).

$$r = \begin{bmatrix} \text{Ant}_1 & \text{Ant}_2 & \cdots & \cdots & \text{Ant}_{n-1} & \text{Ant}_n \\ 1 & 1 & \cdots & \cdots & 1 & 1 \\ 1 & 1 & \cdots & \cdots & 1 & 1 \\ \cdots & \cdots & \cdots & \cdots & \cdots & \cdots \\ \cdots & \cdots & \cdots & \cdots & \cdots & \cdots \\ 1 & 1 & \cdots & \cdots & 1 & 1 \\ 1 & 1 & \cdots & \cdots & 1 & 1 \end{bmatrix} \begin{matrix} \\ Dimention_1 \\ Dimention_2 \\ \\ \\ Dimention_{n-1} \\ Dimention_n \end{matrix}$$

*Figure 25 Current view of (10,30) matrix of ones.*

Initializing (10,30) matrix array to calculate simple random walk for the ants as shown in Figure 26, Figure 27.

```
r = simple_random_walk(rate_of_change)
```

*Figure 26 Initializing random number between [-1,1].*

$$r = \begin{bmatrix} \text{Ant}_1 & \text{Ant}_2 & & & \text{Ant}_{n-1} & \text{Ant}_n \\ [-1,1] & [-1,1] & \cdots & \cdots & [-1,1] & [-1,1] \\ [-1,1] & [-1,1] & \cdots & \cdots & [-1,1] & [-1,1] \\ \cdots & \cdots & \cdots & \cdots & \cdots & \cdots \\ \cdots & \cdots & \cdots & \cdots & \cdots & \cdots \\ [-1,1] & [-1,1] & \cdots & \cdots & [-1,1] & [-1,1] \\ [-1,1] & [-1,1] & \cdots & \cdots & [-1,1] & [-1,1] \end{bmatrix} \begin{matrix} \\ Dimention_1 \\ Dimention_2 \\ \\ \\ Dimention_{n-1} \\ Dimention_n \end{matrix}$$

*Figure 27 Current view of random walk [-1,1] with the shape of (10,30).*

After applying all those steps, the algorithm will pass through three conditions:

- If the current ant is equal to the best ant use equation (3).
- If the current ant is equal to the previous use equation (4).
- If the above points aren't satisfied use equation (5).

Those steps are stated in section 4 in a more detailed way.

Then the process will pass through three conditions to calculate all the processes as shown in Figure 28.

```
Algorithm 2. Pseudocode of VANA algorithm for minimizing the problem
    For a turn-to-range (turn)
        worker_ants ← np.random.uniform(lb,ub,(dim,no_of_agents))
        previous_x ← np.copy(worker_ants)
        Initalize best_ant_index,best_ant_fitness ← to empty array
        For itteration to range(max_itteration)
            rate_of_change ← create (10,30) size of the matrix of zeros
            worker_ant_fitness ← find the fitness of worker ant
            previous_fitness ← find the fitness of the previous ant
            best_ant_index, best_ant_fitness ← stores index of best fitness and stores best ant fitness
            best_ant ← store best ant by ant index
            distance_from_best_ant ← calculate the distance of all ants from the best ant
            temp_xs ← store worker ant in the temporary ant
            r ← simple random walk between [-1,1]
            If worker_ants == best_ant Then
                calculate equation (3)
            EndIf
            If worker_ants == previous_x Then
                calculate equation (4)
            EndIf
            If np.logical_not(cond1 | cond2) Then
                calculate equation (2,5,6,7)
            EndIf
            temp_fitness ← store fitness ant to temporary fitness
            If temp_fitness < worker_ant_fintness Then
                Move to accept
            EndIf
            Else
                maintain the current position
        EndFor
    Endfor
```

*Figure 28 VANA algorithm pseudocode.*

As shown in Figure 3 and Figure 28 ANA algorithm performs slowly due to the number of iterations after modifying the algorithm to vectorization less number of iterations are used as in Figure 4 and Figure 5 initializing random number for ants reduced from 300 iterations to a matrix of size (dimensions, agents), also for copying current worker ant to previous worker ant iteration also neglected as shown in Figure 7, for creating rate of change also 300 iteration are neglected. All the functionality of ANA algorithms, which are performed with iteration are converted to matrices which helps to improve the algorithm's performance and speed. The whole pseudocode of VANA which stands for vectorized ANA algorithm is stated in Figure 28. And the Flowchart of the VANA algorithm is stated in Figure 29. Modifying the ANA algorithm to vectorized method improved the speed of the algorithm as Table 4 and Table 5 indicate that the algorithms perform much faster mainly in test functions of F9, F12, F15, F17, and F18, also in CEC01, which was taken around 6 hours to execute the equation but it takes only 1.5 minutes to execute the equation also for CEC06 was taking around 3 hours with vectorization takes 30 seconds which it concludes that the speed of the algorithm was mainly from the iteration which can be improved by vectorization all comparison of ANA and VANA are stated in section 5 including Figure 30, Figure 31, Figure 32 and Figure 33.

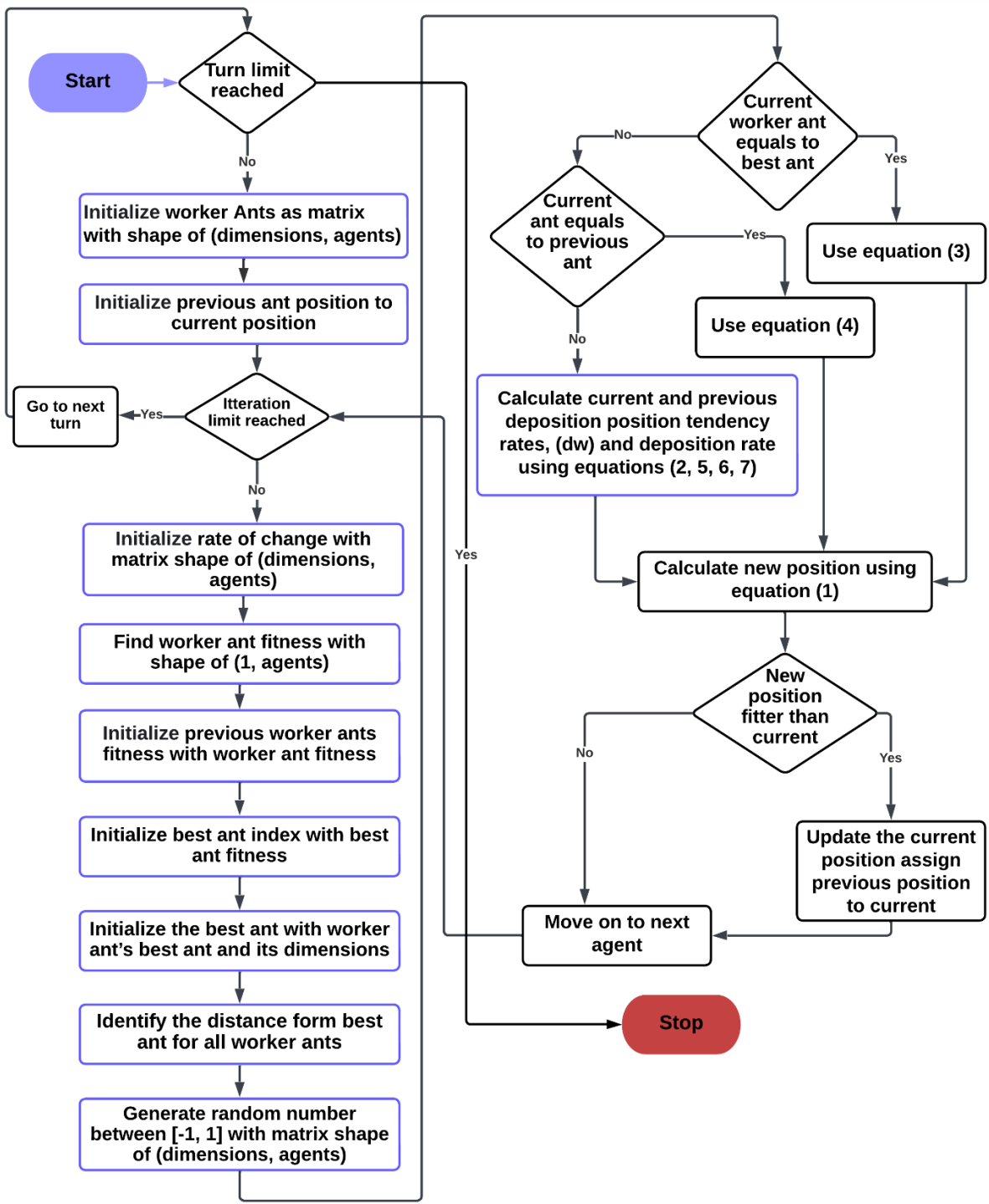

*Figure 29 Flowchart of the VANA algorithm.*

Table 4 Time in seconds for F1 to F18 test functions.

| Test Function | Execution Time | |
|---|---|---|
| | **ANA** Algorithm | **VANA** Algorithm |
| F1 | 32 Seconds | 9.5 Seconds |
| F2 | 57 Seconds | 9.7 Seconds |
| F3 | 52 Seconds | 9.6 Seconds |
| F4 | 28 Seconds | 25.9 Seconds |
| F5 | 26 Seconds | 12.4 Seconds |
| F7 | 26 Seconds | 11 Seconds |
| F9 | **106 Seconds** | **11.1 Seconds** |
| F10 | 16 Seconds | 11.7 Seconds |
| F11 | **168 Seconds** | **12 Seconds** |
| F12 | **140 Seconds** | **18 Seconds** |
| F13 | 6 Seconds | 4.9 Seconds |
| F14 | 39 Seconds | 6.7 Seconds |
| F15 | **151 Seconds** | **11.6 Seconds** |
| F16 | 79 Seconds | 15.5 Seconds |
| F17 | **81 Seconds** | **15 Seconds** |
| F18 | **96 Seconds** | **17.5 Seconds** |

Table 5 Time in seconds for CEC01 to CEC10 test functions.

| Test Function | Execution Time | |
|---|---|---|
| | **ANA** Algorithm | **VANA** Algorithm |
| **CEC01** | **23705 Seconds** | **94 Seconds** |
| **CEC02** | 58 Seconds | 18.4 Seconds |
| **CEC03** | 511 Seconds | 458 Seconds |
| **CEC04** | 208 Seconds | 12.9 Seconds |
| **CEC05** | 337 Seconds | 12.7 Seconds |
| **CEC06** | **10606 Seconds** | **22.9 Seconds** |
| **CEC07** | 341 Seconds | 22.9 Seconds |
| **CEC08** | 266 Seconds | 17.8 Seconds |
| **CEC09** | 148 Seconds | 12.9 Seconds |
| **CEC10** | 17 Seconds | 12.5 Seconds |

## 6. ANA VS VANA Speed Comparison

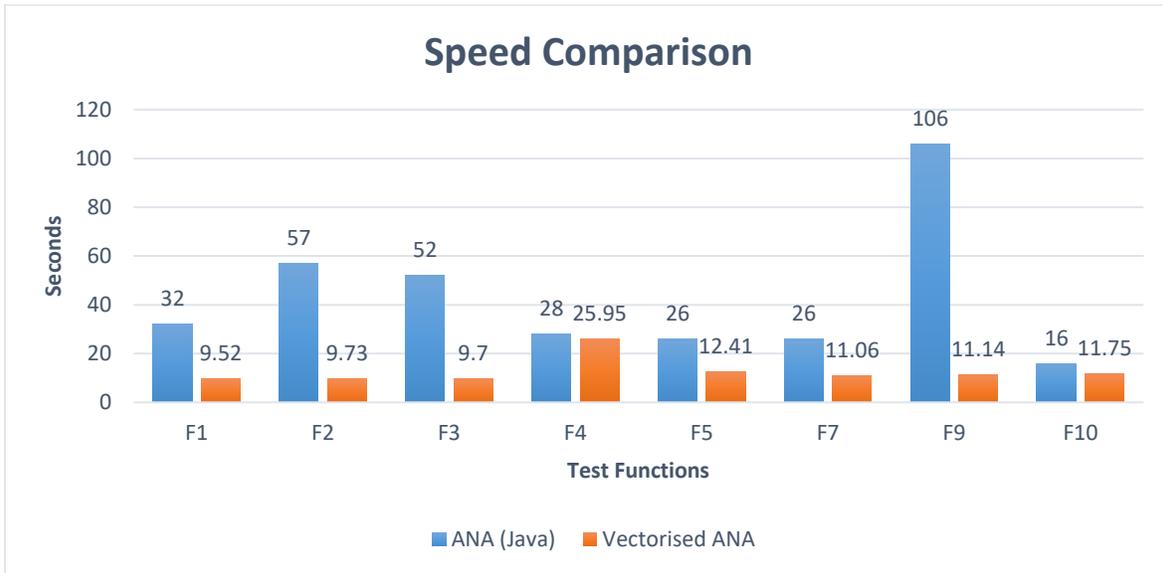

Figure 30 Comparison between VANA and ANA from F1 to F10 in seconds.

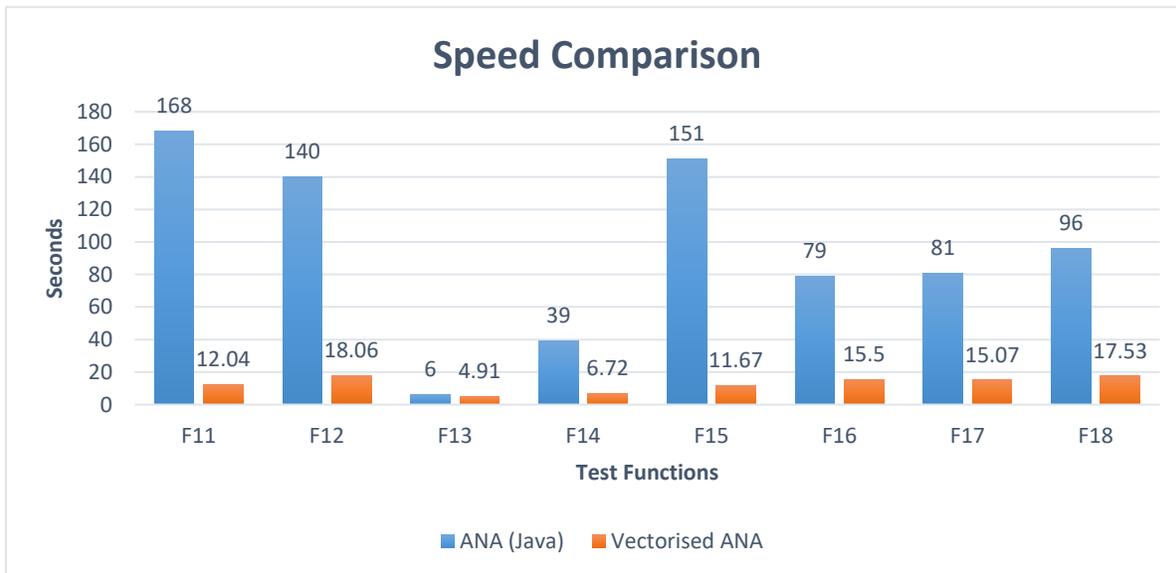

Figure 31 Comparison ANA VS VANA from F11 to F18 in seconds.

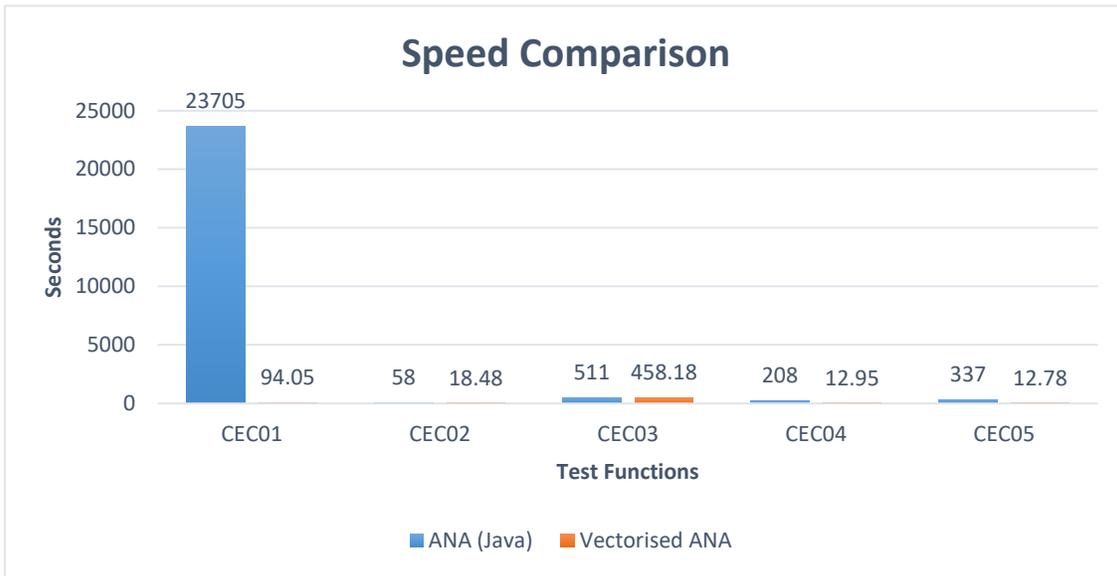

*Figure 32 Comparison of ANA VS VANA from CEC01 to CEC05 in seconds.*

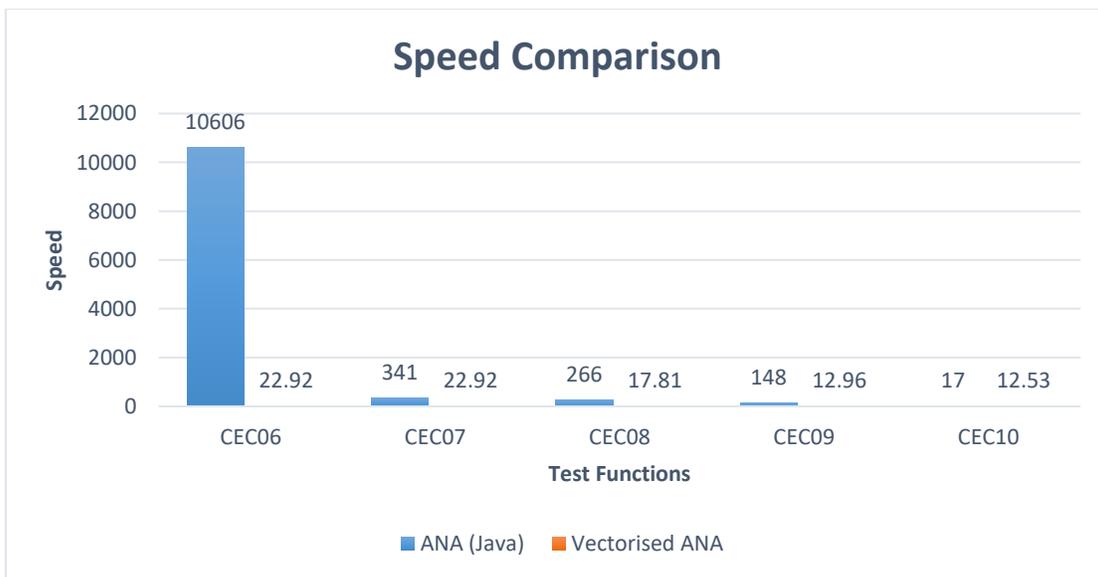

*Figure 33 Comparison of ANA VS VANA from CEC06 to CEC10 in seconds.*

## 7. CONCLUSION

This paper introduces a new method for using optimization techniques to improve the algorithm's speed and performance, since the metaheuristic algorithm uses test functions that perform slowly in execution time, the vectorized method it improves the speed of the algorithm. Vectorization is a process of executing multiple statements at a time which enhances the performance in speed of the algorithm. This method was straightforwardly used to change the algorithm to vectorization to implement slow test functions, reduce processing time and loads of processing and reduce delay. Overall vectorization of algorithms is strong and well-suited for metaheuristic algorithms to solve the problems of optimization in a shorter period. Vectorization can be applied to other algorithms especially ones with real-world application that requires rapid speed.

**Funding:** This research received no specific grant from any funding agency in the public, commercial, or not-for-profit sectors.

**Data Availability:** The data used to support the findings of this study are available from the corresponding author upon request.

**Acknowledgment:** The authors would like to express their appreciation to the University of Kurdistan Hewler.

**Conflicts of Interest** The authors have no conflicts of interest to declare.